\documentclass[10pt,twocolumn]{article}

\usepackage[a4paper,margin=0.7in]{geometry}
\usepackage[T1]{fontenc}
\usepackage{lmodern}

\usepackage{graphicx}
\usepackage{booktabs}
\usepackage{multirow}
\usepackage{dblfloatfix}
\usepackage{amsmath}
\usepackage{amssymb}
\usepackage{microtype}
\usepackage{cite}
\usepackage[hidelinks]{hyperref}
\usepackage{xcolor}
\usepackage{titling}

\newcommand{\Description}[1]{}

\pretitle{\centering\Large\bfseries}
\posttitle{\par\vskip 0.5em}

\title{
\textbf{Learning to Fly: Stable Vision-Guided UAV Servoing with\\
Compact Target-Centric Cues and Reinforcement Learning}
}

\author{
Saurbh Singh Jamwal, Nived Chebrolu\\
Department of Computer Science and Engineering\\
Indian Institute of Technology Bombay\\
\texttt{\{saurbh,nived\}@cse.iitb.ac.in}
}

\date{}

\begin{document}
\raggedbottom

\maketitle

\begin{abstract}

Vision-guided reinforcement learning for Unmanned Aerial Vehicles (UAVs) remains challenging due to unstable policy optimisation, aggressive exploration, and the cost of high-dimensional visual perception. In this work, we investigate long-horizon UAV visual servoing using compact target-centric cues combined with low-dimensional sensor measurements. Rather than learning directly from RGB images, lightweight target segmentation provides image-space offsets and relative depth, which are combined with quadrotor velocity and projected-gravity measurements into a compact 12D policy observation. We compare Direct PPO with three matched-budget curriculum strategies: a Visual curriculum that progressively expands target placement difficulty, a Dynamics curriculum that gradually relaxes action constraints and smoothing, and a Joint curriculum that combines both progressions. All strategies reach comparable nominal performance, with complementary advantages across tracking metrics. Observation ablations show that proprioceptive measurements are critical for stable flight and image-space cues for target alignment, while explicit depth is not necessary for strong performance in the evaluated setting. Against tuned classical visual-servo controllers, learned policies show greater robustness to strong control and visual perturbations, while the Visual curriculum exhibits the smallest degradation under unseen target motion. Overall, the results demonstrate that compact target-centric representations can support robust long-horizon aerial visual servoing and that visual curriculum training can improve robustness to dynamic distribution shifts despite limited gains in nominal performance.

\end{abstract}

\vspace{0.5em}

\section{Introduction}

Unmanned aerial vehicles (UAVs) have become increasingly important for applications such as remote sensing, aerial surveillance, search and rescue operations, civil infrastructure inspection, and precision agriculture~\cite{shakhatreh2019uavsurvey}. Many of these tasks require UAVs to operate robustly in dynamic and uncertain environments while maintaining stable control under continuously changing control demands and visual conditions. However, achieving reliable autonomous flight remains challenging due to the highly nonlinear, coupled, and underactuated dynamics of quadrotor systems.

Classical UAV control approaches based on PID control, Linear Quadratic Regulators (LQR), and Model Predictive Control (MPC) have demonstrated strong performance for the stabilisation and trajectory tracking under carefully modeled system dynamics~\cite{lopez2023pid,mellinger2011minimum,faessler2017differential}. However, these methods often require extensive system identification, expert tuning, accurate state estimation, and carefully designed control pipelines, which can become increasingly difficult to maintain under rapidly changing environments and complex flight conditions~\cite{lopez2023pid}. Reinforcement learning (RL) has therefore emerged as a promising alternative by enabling UAVs to learn control policies directly from interaction and experience. Recent advances in RL-based UAV control have demonstrated aggressive maneuvering, agile flight, and autonomous policy learning under complex nonlinear dynamics~\cite{hwangbo2017control,kaufmann2020deep,wang2024constrained}.

At the same time, vision-guided UAV control has gained attention for enabling aerial systems to operate using onboard visual feedback without relying heavily on external localisation or mapping infrastructure. Recent works have explored end-to-end visual navigation, obstacle avoidance, autonomous target following, and visual servoing using deep learning and reinforcement learning frameworks. Despite these advances, stable learning for vision-guided UAV control remains challenging due to unstable policy optimisation, high-dimensional visual observations, and sensitivity to challenging operating conditions and observation perturbations.

In this work, we investigate a reinforcement learning framework for stable long-horizon aerial visual servoing using compact target-centric cues and low-dimensional sensor measurements. Rather than learning directly from high-dimensional image streams, the proposed framework extracts image-space target offsets and relative depth from lightweight target segmentation. These compact visual cues are combined with quadrotor velocity and projected-gravity measurements to form a 12D policy observation. Within this representation, we perform a controlled study of Direct PPO and visual, dynamics, and joint curriculum strategies under matched optimisation budgets. Beyond nominal performance, we investigate which components of the compact observation are necessary for successful control, compare learned policies against classical 2D and 3D visual-servo controllers, and evaluate robustness to control and perception perturbations as well as previously unseen target motion. This allows us to study not only whether curriculum learning facilitates optimisation, but also whether the resulting policies differ in robustness and generalisation despite similar nominal performance.

The contributions of this work are summarised as follows:

\begin{itemize}

\item We develop a lightweight vision-guided UAV control framework that represents the visual servoing task using compact target-centric image offsets and relative depth together with low-dimensional quadrotor measurements, forming a 12D policy observation without requiring high-dimensional visual feature extraction.

\item We conduct a controlled comparison of Direct PPO and Visual, Dynamics, and Joint curriculum strategies under matched optimisation budgets, showing that the approaches reach comparable long-horizon nominal performance while exhibiting complementary advantages across tracking and control metrics.

\item Through systematic observation ablations, we analyse the contribution of visual and proprioceptive information to the learned controller, showing that proprioception is critical for stable flight and image-space target cues for servo alignment, while explicit target depth is not necessary for strong performance in the evaluated operating range.

\item We benchmark the learned policies against tuned 2D and 3D classical visual-servo controllers under control noise, visual corruption, dropout, and previously unseen target motion. The results reveal greater robustness of the learned policies under strong perturbations and show that visual curriculum training reduces degradation under dynamic target motion despite comparable nominal performance.

\end{itemize}

\section{Related Work}

\paragraph{Classical UAV Control}

Classical quadrotor control has traditionally relied on PID controllers, trajectory optimisation, and differential flatness-based planning. PID-based approaches remain widely used for UAV stabilisation and trajectory tracking due to their simplicity and effectiveness~\cite{lopez2023pid}. Trajectory generation methods such as minimum-snap optimisation~\cite{mellinger2011minimum} and differential flatness-based tracking under aerodynamic effects~\cite{faessler2017differential} have enabled accurate and dynamically feasible quadrotor flight. More recent works have explored integrating learned dynamics models with classical control frameworks to improve agile flight performance~\cite{saviolo2023learning}. Despite their strong performance, these approaches generally depend on accurate system modeling, expert tuning, and carefully designed control pipelines, motivating the exploration of learning-based UAV control strategies.

\paragraph{Reinforcement Learning for UAV Control}

Reinforcement learning (RL) has increasingly emerged as an alternative to classical model-based UAV control by enabling quadrotors to learn control policies directly from interaction rather than relying entirely on accurate system modeling and hand-tuned controllers. Early works demonstrated that deep RL could successfully stabilise quadrotor flight and learn low-level motor control policies capable of real-world deployment~\cite{hwangbo2017control}. These developments marked a significant shift toward data-driven policy learning for highly nonlinear and underactuated aerial systems.

As RL methods matured, research focus expanded from basic stabilisation toward aggressive maneuvering, agile flight, and robust autonomous control. Kaufmann et al.~\cite{kaufmann2020deep} demonstrated high-speed agile quadrotor flight through efficient simulation-based training, while more recent works have explored robust and constrained UAV control under challenging flight conditions~\cite{wang2024constrained}. In parallel, developmental and curriculum-inspired learning strategies have been investigated to improve training stability and policy acquisition for complex UAV dynamics by progressively increasing task difficulty during learning~\cite{deshpande2020developmental}.

Despite these advances, stable long-horizon learning, sample-efficient training, and robust recovery behavior remain challenging for RL-based UAV control systems, particularly in vision-guided settings where noisy observations, unstable exploration dynamics, and difficult control conditions can significantly affect policy performance.

\paragraph{Vision-Based UAV Navigation and Control}

Vision-based UAV control has evolved from classical visual servoing approaches toward modern deep learning and reinforcement learning-based visuomotor policies. Early visual servoing methods utilised image-space feedback to directly control robotic motion through visual error minimisation~\cite{chaumette2006visual}. These approaches established the foundation for target-centric aerial control using visual observations rather than explicit global localisation or trajectory planning. Early UAV tracking systems further explored vision-based target following and GPS-denied navigation using handcrafted perception and control pipelines~\cite{pestana2013vision}.

With the emergence of deep learning, research focus shifted toward end-to-end visual control policies learned directly from image observations. Gandhi et al.~\cite{gandhi2017learning} demonstrated self-supervised visual navigation for UAVs, while DroNet~\cite{loquercio2018dronet} showed that compact convolutional architectures could enable efficient real-time aerial navigation and obstacle avoidance from monocular visual input. Subsequent works further explored agile autonomous flight and reinforcement learning-based visual UAV training under complex navigation environments~\cite{kaufmann2018deep,krishnan2021airlearning}.

More recent studies have investigated reinforcement learning for visual servoing and target-centric UAV control. Jin et al.~\cite{jin2021policy} explored policy-based deep reinforcement learning for visual servoing under visibility constraints, while Fu et al.~\cite{fu2023deep} investigated RL-based UAV visual servoing with field-of-view constraints. Additional works have examined autonomous visual target following and multi-UAV tracking in cluttered environments~\cite{chen2023quadcopter,hung2022image}.

\begin{figure*}[t]
  \centering
  \includegraphics[width=\textwidth]{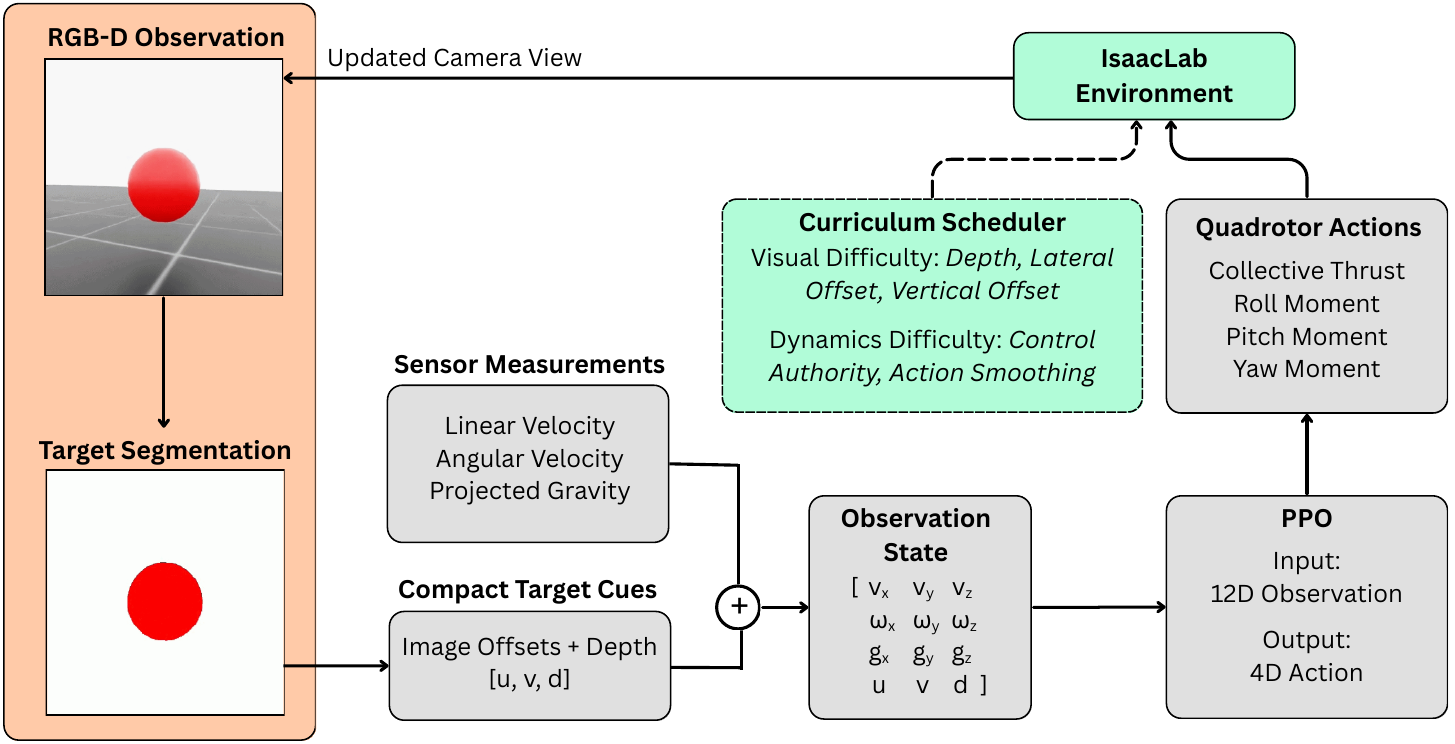}
  \caption{Overview of the proposed vision-guided UAV visual servoing framework. The RGB-D observation is processed through target segmentation to extract compact image-offset and depth cues, which are combined with sensor measurements into a 12D observation state. A PPO policy maps this state to collective thrust and body-moment commands, while the curriculum scheduler adjusts the IsaacLab training difficulty.}
  \Description{System overview of the proposed UAV visual servo reinforcement learning pipeline showing RGB-D observation, target segmentation, compact cue extraction, 12D observation construction, PPO policy inference, curriculum scheduling, and quadrotor action execution.}
  \label{fig:pipeline}
\end{figure*}

Despite these advances, vision-guided UAV learning commonly focuses on obstacle avoidance, navigation, or visuomotor policies operating on comparatively rich visual representations. For visual servoing, however, the control objective is naturally expressed through target-relative geometry. This motivates asking whether high-dimensional visual features are necessary once task-relevant information has been extracted from the image.

Our work therefore studies long-horizon aerial visual servoing through a compact control-oriented interface consisting of normalised image-space target offsets and relative depth, combined with low-dimensional quadrotor measurements. The perception module is used only to produce these target-centric cues, separating the visual representation from the downstream control policy. This allows us to directly examine whether lightweight geometric feedback is sufficient for stable reinforcement learning-based aerial servoing, and which components of this representation are actually required for successful control.

\paragraph{Curriculum Learning for UAVs}

Curriculum learning progressively increases task difficulty to improve
reinforcement learning stability and exploration. Bengio et al.~\cite{bengio2009curriculum} introduced the general framework, while subsequent work explored reverse curriculum~\cite{florensa2017reverse},
adaptive teacher-student strategies~\cite{matiisen2019teacher}, and progressive environment randomisation~\cite{akkaya2019solving}. These studies show that structured task progression can improve exploration and convergence in complex reinforcement learning problems. For UAVs, this is particularly relevant because nonlinear dynamics and aggressive exploration can hinder stable policy
optimisation. Deshpande et al.~\cite{deshpande2020developmental} studied developmental UAV control, while Sun et al.~\cite{sun2026curriculum} applied curriculum learning to progressively challenging agile navigation.

Building on this motivation, we study curriculum learning specifically for compact target-centric aerial visual servoing. Rather than evaluating a single curriculum against direct training, we separately vary visual difficulty, control difficulty, and their joint progression while keeping the policy, reward, optimisation budget, and final task fixed. This controlled design asks whether curriculum structure changes only optimisation and convergence, or also the behavior of the resulting policies after nominal performance becomes comparable. We therefore evaluate the learned controllers beyond the training distribution through unseen target motion and sensing and control perturbations, and compare them with classical visual-servo baselines.

\section{Methodology}

Figure~\ref{fig:pipeline} illustrates the overall visual servo reinforcement learning framework used in this work.

\subsection{Problem Formulation and Compact Observation Design}
\label{sec:compact_observation}

We address long-horizon aerial visual servoing for quadrotor control, where the objective is to maintain image-space target alignment and target-relative depth while preserving stable flight. Rather than learning directly from high-dimensional RGB observations, we represent the visual servoing task using the compact target-centric cue
\[
\mathbf{z}_t=[u_t,v_t,d_t],
\]
where $u_t$ and $v_t$ denote the normalised horizontal and vertical target offsets and $d_t$ denotes the normalised target-relative depth.

As illustrated in Figure~\ref{fig:pipeline}, a target segmentation module first extracts the target mask from the RGB observation. In this work, we use lightweight color-based segmentation in IsaacLab for experimental control; the purpose is not to benchmark segmentation algorithms, but to study control using compact mask-derived cues. Let $(x_t,y_t)$ denote the target centroid, $(c_x,c_y)$ the image center, and $W$ and $H$ the image dimensions. The image-space offsets are
\[
u_t=\frac{x_t-c_x}{W}, \qquad
v_t=\frac{y_t-c_y}{H}.
\]
The depth component $d_t$ is normalised relative to the desired operating range. In IsaacLab, relative depth is obtained from the simulator for controlled evaluation; however, it enters the policy only as a scalar cue and can equivalently be supplied by RGB-D sensing, stereo, monocular depth estimation, or learned depth modules.

The resulting representation directly captures horizontal alignment, vertical alignment, and target-relative distance while avoiding high-dimensional visual feature extraction. The perception module is therefore interchangeable: alternative segmentation models, detectors, trackers, or depth estimators can be used provided they produce the same compact target-centric interface. Finally, $\mathbf{z}_t$ is combined with low-dimensional quadrotor measurements to form the policy observation described in Section~\ref{sec:rl_framework}, from which the learned policy generates continuous control commands for closed-loop visual tracking and flight stabilisation.

\subsection{Reinforcement Learning Framework}
\label{sec:rl_framework}

We formulate aerial visual servoing as a continuous-control problem trained using Proximal Policy Optimisation (PPO). At each timestep $t$, the compact target cue $\mathbf{z}_t$ is combined with body-frame linear velocity $\mathbf{v}_t$, angular velocity $\boldsymbol{\omega}_t$, and projected gravity $\mathbf{g}_t$ to form the 12-dimensional policy observation
\[
\mathbf{o}_t =
[\mathbf{v}_t,\boldsymbol{\omega}_t,\mathbf{g}_t,\mathbf{z}_t],
\qquad
\mathbf{z}_t=[u_t,v_t,d_t].
\]

The policy $\pi_\theta$ outputs the four-dimensional continuous action
\[
\mathbf{a}_t =
[a_t^{T},a_t^{\tau_x},a_t^{\tau_y},a_t^{\tau_z}],
\]
corresponding to collective thrust and body-frame roll, pitch, and yaw moments. The normalised actions are scaled to force and torque commands and applied directly to the quadrotor body in IsaacLab. PPO optimises the expected discounted return
\[
J(\theta)=
\mathbb{E}_{\pi_\theta}
\left[
\sum_{t=0}^{T}\gamma^t r_t
\right],
\]
where $r_t$ is the timestep reward and $\gamma$ is the discount factor. Training is performed using parallel IsaacLab environments.

\subsection{Reward Design}

The reward encourages long-horizon visual servoing by balancing target alignment, depth regulation, visibility, and flight stability. It is defined as
\[
r_t =
\lambda_c r_t^{\text{center}}
+\lambda_d r_t^{\text{depth}}
+\lambda_v r_t^{\text{visible}}
-\lambda_\omega r_t^{\text{ang}}
-\lambda_h r_t^{\text{height}},
\]
where the terms respectively capture image-space target alignment, target-relative depth regulation, target visibility, angular-motion regularisation, and altitude stabilisation. The same reward formulation and weights are used for Direct PPO and all curriculum strategies, ensuring that their comparison isolates differences in training progression.

For evaluation, a timestep is considered successful when the target is visible and both the image-space center error and depth error are below fixed thresholds.

\subsection{Curriculum Learning Strategy}

We compare Direct PPO with Visual, Dynamics, and Joint curricula to study the effect of progressive task difficulty. All methods use identical policy architecture, reward, PPO hyperparameters, and optimisation budget; only the difficulty schedule differs. Curriculum progress is represented by $p\in[0,1]$, progressing from the easiest ($p=0$) to the final task ($p=1$) according to a fixed schedule. For any curriculum parameter $q$, we use linear interpolation
\[
q(p)=q_{\mathrm{easy}}+p(q_{\mathrm{final}}-q_{\mathrm{easy}}).
\]
Direct PPO uses the final task setting ($p=1$) throughout training.

\paragraph{Visual Curriculum.}
Target placement difficulty is progressively increased while keeping the dynamics setting fixed. The maximum target depth, lateral displacement, and vertical displacement are interpolated from $2.0$ to $5.0$~m, $0.20$ to $1.50$~m, and $0.10$ to $1.00$~m, respectively. Thus, training progresses from nearby, weakly displaced targets to the full target distribution.

\paragraph{Dynamics Curriculum.}
Control difficulty is increased while keeping the visual distribution fixed. The action limit $a_{\max}$ progresses from $0.35$ to $1.00$, and the body-moment scale $m$ from $0.0015$ to $0.0030$, using the interpolation above. To suppress aggressive early exploration, actions are additionally filtered as
\[
\tilde{\mathbf{a}}_t=\alpha(p)\tilde{\mathbf{a}}_{t-1}
+\left(1-\alpha(p)\right)\mathbf{a}_t,
\]
where $\alpha$ decreases from $0.50$ to $0.10$. Hence, control authority increases and smoothing decreases as training progresses.

\paragraph{Joint Curriculum.}
The Joint curriculum simultaneously applies the Visual and Dynamics progressions using the same $p$. At $p=1$, all curriculum variants recover the same visual distribution and control setting used by Direct PPO, ensuring that they differ only in their progression through task difficulty rather than the final task.

\subsection{Classical Visual-Servo Baselines}
\label{sec:classical_servo}

To provide non-learning references, we implement two deterministic classical visual-servo baselines, denoted Classical Servo 2D and Classical Servo 3D, using the same compact target observations and onboard quadrotor state available to the learned policies. Both controllers produce the same normalised four-dimensional action vector as PPO, consisting of collective thrust and roll, pitch, and yaw moment commands. No target world position is used.

Let $u_t$ and $v_t$ denote the normalised horizontal and vertical image offsets and let
\[
e_t^d = d_t-d^\star
\]
denote the depth error relative to the desired target distance $d^\star$. Finite-difference derivatives $\dot u_t$, $\dot v_t$, and $\dot e_t^d$ are computed from consecutive observations. Horizontal image error controls yaw through
\[
a_t^{\tau_z} =
s_u\left(K_{p,u}u_t+K_{d,u}\dot u_t\right)
-K_{d,\omega_z}\omega_{z,t},
\]
while vertical image error modifies collective thrust according to
\[
a_t^{T} =
a_{\mathrm{hover}}
+s_v\left(K_{p,v}v_t+K_{d,v}\dot v_t\right)
-K_{d,v_z}v_{z,t},
\]
where $s_u$ and $s_v$ account for camera/body sign conventions and $a_{\mathrm{hover}}=2/\rho-1$ is the normalised hover command for thrust-to-weight ratio $\rho$.

Roll and pitch are stabilised using projected gravity and body angular rates:
\[
a_t^{\tau_x} =
s_r K_{p,r}g_{y,t}
-K_{d,r}\omega_{x,t},
\]
\[
a_t^{\tau_y} =
s_p K_{p,p}g_{x,t}
-K_{d,p}\omega_{y,t}
+a_t^{\mathrm{vis},p}.
\]

For the Classical Servo 2D, $a_t^{\mathrm{vis},p}=0$, so the controller performs image-space centering without explicitly regulating target depth. For the Classical Servo 3D, depth error additionally contributes a pitch command
\[
a_t^{\mathrm{vis},p}
=
s_d\left(K_{p,d}e_t^d+K_{d,d}\dot e_t^d\right).
\]

All commands are clipped to fixed safety limits. When the target is not considered visible, the outer visual-servo terms are suppressed and only the inner stabilisation terms remain active. Controller gains are tuned under the nominal static-target setting and then kept fixed for all moving-target and robustness evaluations.

\subsection{Robustness Evaluation Protocol}

To evaluate robustness under degraded sensing and control conditions, we apply perturbations only at evaluation time while keeping the trained policies and controller parameters fixed. We consider three perturbations: action noise, visual observation noise, and intermittent visual dropout. Action noise adds zero-mean Gaussian noise to the normalised control commands, visual noise applies Gaussian perturbations to the compact target-centric cues, and dropout temporarily removes valid visual observations. All perturbations use level $0.2$: the Gaussian perturbations have standard deviation $0.2$ in their respective normalised spaces, while visual dropout occurs with probability $0.2$ per timestep. No additional training or controller retuning is performed.

We evaluate Direct PPO, Visual Curriculum PPO (selected for its strong task-success performance), and the classical visual-servo baselines under the same perturbation conditions. Performance is measured using cumulative reward, center error, depth error, visibility rate, successful-step rate, and survival rate. Survival rate denotes the fraction of episodes reaching the maximum horizon without premature termination. This protocol therefore measures zero-shot robustness to sensing and control degradation rather than robustness acquired through perturbation-aware training.

\section{Experiments}

\subsection{Experimental Setup}

All experiments are conducted in the IsaacLab physics-based simulator using a quadrotor with a forward-facing RGB camera and a target-relative depth cue provided by the simulator. Target masks are obtained using the modular segmentation block described in Section~\ref{sec:compact_observation}, and the resulting target-centric cues are combined with quadrotor sensor measurements as described in Section~\ref{sec:rl_framework}.

Policies are trained using PPO with continuous collective-thrust and body-moment commands. Unless otherwise stated, evaluation uses 200 episodes per run and five independent runs. We report episodic reward, center error, depth error, successful-step rate, visibility rate, and survival rate. Successful-step rate is the fraction of evaluation steps for which the target is visible and both center and depth errors satisfy the predefined tracking thresholds. Survival rate is the fraction of episodes that terminate by reaching the time limit rather than through flight failure.

We compare Direct PPO with Visual Curriculum PPO, Dynamics Curriculum PPO, and Joint Curriculum PPO under matched optimisation budgets. For generalisation and robustness experiments, we additionally compare against Classical Servo 2D and Classical Servo 3D. All learned policies remain fixed during evaluation, and classical-controller gains are not retuned for individual test conditions.

\subsection{Equal-Budget Curriculum Analysis}

We first isolate the effect of curriculum structure by comparing Direct PPO, Visual Curriculum PPO, Dynamics Curriculum PPO, and Joint Curriculum PPO under identical PPO hyperparameters, environment counts, training iterations, and evaluation protocols across five independent random seeds.

\begin{figure*}[t]
    \centering
    \includegraphics[width=0.93\textwidth]{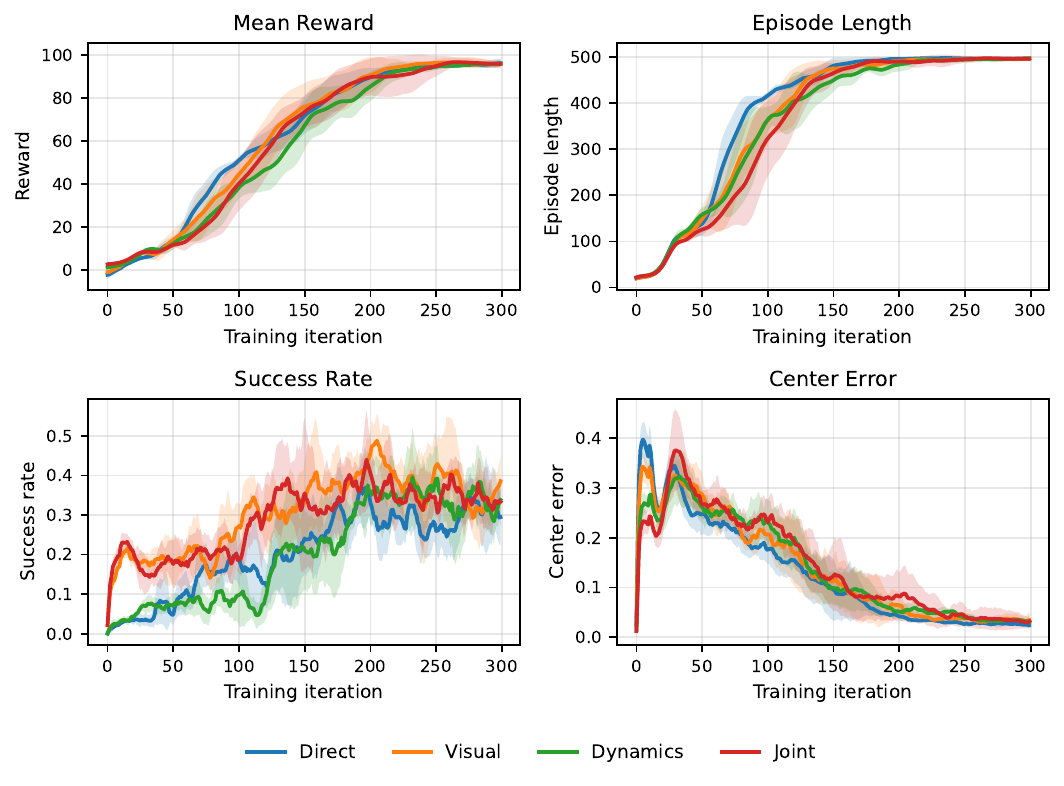}
    \caption{Equal-budget training comparison averaged across five random seeds. The panels show (\textbf{a}) episodic reward, (\textbf{b}) episode length, (\textbf{c}) successful-step rate, and (\textbf{d}) center error. Shaded regions indicate one standard deviation.}
    \Description{Training curves comparing Direct PPO, Visual, Dynamics, and Joint curriculum strategies under matched optimisation budgets.}
    \label{fig:week6_training_curves}
\end{figure*}

\begin{table*}[t]
    \caption{Equal-budget comparison under the hardest static evaluation configuration, averaged across five random seeds. Bold indicates the best mean value.}
    \label{tab:curriculum_comparison}
    \centering
    \resizebox{\textwidth}{!}{%
    \begin{tabular}{lccccc}
    \toprule
    \textbf{Method} &
    \textbf{Reward $\uparrow$} &
    \textbf{Ep. Length $\uparrow$} &
    \textbf{Center Error $\downarrow$} &
    \textbf{Depth Error $\downarrow$} &
    \textbf{Success $\uparrow$} \\
    \midrule
    Direct PPO
    & 96.01 $\pm$ 1.89
    & 495.9 $\pm$ 4.7
    & \textbf{0.0233 $\pm$ 0.0077}
    & 1.3925 $\pm$ 0.6671
    & 0.2825 $\pm$ 0.2046 \\

    Visual Curriculum PPO
    & 95.94 $\pm$ 1.41
    & 496.4 $\pm$ 4.9
    & 0.0388 $\pm$ 0.0329
    & \textbf{1.1669 $\pm$ 0.2578}
    & \textbf{0.3835 $\pm$ 0.1306} \\

    Dynamics Curriculum PPO
    & \textbf{96.56 $\pm$ 2.07}
    & \textbf{498.2 $\pm$ 1.6}
    & 0.0257 $\pm$ 0.0113
    & 1.4297 $\pm$ 0.2884
    & 0.3257 $\pm$ 0.1054 \\

    Joint Curriculum PPO
    & 96.35 $\pm$ 2.78
    & 496.9 $\pm$ 2.8
    & 0.0515 $\pm$ 0.0539
    & 1.7296 $\pm$ 0.5923
    & 0.2877 $\pm$ 0.2242 \\
    \bottomrule
    \end{tabular}%
    }
\end{table*}

Figure~\ref{fig:week6_training_curves} and Table~\ref{tab:curriculum_comparison} show that all strategies reach comparable long-horizon performance when optimisation budgets are matched. Dynamics Curriculum PPO obtains the highest reward and episode length, Visual Curriculum PPO achieves the highest successful-step rate and lowest depth error, and Direct PPO produces the lowest center error. Thus, curriculum learning does not universally improve asymptotic performance, but changes the resulting tracking trade-offs. We use Visual Curriculum PPO in subsequent generalisation experiments because it provides the strongest task-success performance among the curriculum variants.

\subsection{Observation Ablation}

We next examine whether the proposed compact observation components contribute meaningfully to control performance. Starting from the full 12D Direct PPO observation, we remove target-relative depth or proprioceptive measurements and additionally evaluate a state-only variant. All variants use the same PPO training and evaluation protocol.

\begin{table*}[t]
    \caption{Observation ablation under the hard evaluation setting. Results are mean $\pm$ standard deviation across five seeds. The variants of PPO isolate the roles of target-relative depth, proprioception, and target-centric visual cues.}
    \label{tab:observation_ablation}
    \centering
    \small
    \setlength{\tabcolsep}{4pt}
    \begin{tabular}{lcccccc}
    \toprule
    \textbf{Observation} &
    \textbf{Dim.} &
    \textbf{Reward $\uparrow$} &
    \textbf{Center Error $\downarrow$} &
    \textbf{Depth Error $\downarrow$} &
    \textbf{Success $\uparrow$} &
    \textbf{Survival $\uparrow$} \\
    \midrule

    Full
    & 12
    & 89.09 $\pm$ 3.22
    & 0.0230 $\pm$ 0.0032
    & 0.9779 $\pm$ 0.1060
    & 0.4719 $\pm$ 0.0371
    & 97.1\% \\

    No Depth
    & 11
    & \textbf{94.17 $\pm$ 3.72}
    & \textbf{0.0182 $\pm$ 0.0052}
    & \textbf{0.7986 $\pm$ 0.1100}
    & \textbf{0.5318 $\pm$ 0.0605}
    & \textbf{98.7\%} \\

    No Proprioception
    & 3
    & 2.87 $\pm$ 1.09
    & 0.2649 $\pm$ 0.0234
    & 0.8022 $\pm$ 0.0846
    & 0.3674 $\pm$ 0.0516
    & 4.4\% \\

    State Only
    & 9
    & 58.01 $\pm$ 4.30
    & 0.2649 $\pm$ 0.0389
    & 0.8494 $\pm$ 0.3045
    & 0.3131 $\pm$ 0.0937
    & \textbf{100.0\%} \\
    \bottomrule
    \end{tabular}
\end{table*}

Table~\ref{tab:observation_ablation} reveals distinct roles for the observation components. Removing proprioception causes severe flight instability, reducing survival from $97.1\%$ to $4.4\%$, whereas the state-only policy achieves $100\%$ survival but substantially worse target centering and successful-step rate. Thus, proprioception primarily supports flight stability, while target-centric visual cues are necessary for accurate servoing. Surprisingly, removing explicit depth does not degrade performance; the 11D no-depth policy achieves the highest reward, lowest center error, and highest successful-step rate. This suggests that explicit depth is not essential in the evaluated operating regime, although this result should not be interpreted as establishing depth as unnecessary under broader target-motion or depth-variation conditions.

\subsection{Generalisation to Unseen Target Motion}

We next evaluate zero-shot generalisation beyond the static target distribution used during training. Direct PPO and Visual Curriculum PPO are trained exclusively on static targets, while the classical controllers are tuned only under static conditions. We introduce a fast lateral sinusoid ($0.25$~Hz, $0.60$~m amplitude) and a hard helical trajectory ($0.25$~Hz; $0.60$, $0.45$, and $0.35$~m lateral, depth, and vertical amplitudes, respectively), without retraining or gain retuning.

\begin{table*}[t]
  \caption{Zero-shot target-motion generalisation. $\Delta R_{\mathrm{H}}$ is the reward change from static to hard-helix evaluation. Results are mean $\pm$ standard deviation over five runs.}
  \label{tab:moving_target}
  \centering
  \small
  \setlength{\tabcolsep}{4pt}
  \begin{tabular}{lcccccc}
  \toprule
  \textbf{Method} &
  \textbf{Static $R$} &
  \textbf{Lateral $R$} &
  \textbf{Helix $R$} &
  \textbf{$\Delta R_{\mathrm{H}}$} &
  \textbf{Helix Success} &
  \textbf{Helix Survival} \\
  \midrule

  Direct PPO
  & 89.37 $\pm$ 3.26
  & 87.86 $\pm$ 3.71
  & 82.60 $\pm$ 3.91
  & $-7.58\%$
  & 0.4050 $\pm$ 0.0691
  & 97.3\% \\

  Visual Curriculum PPO
  & 87.41 $\pm$ 1.44
  & 88.60 $\pm$ 2.12
  & 83.97 $\pm$ 1.51
  & \textbf{$-$3.94\%}
  & 0.4216 $\pm$ 0.0247
  & \textbf{97.7\%} \\

  Classical Servo 2D
  & 99.37 $\pm$ 1.28
  & 97.29 $\pm$ 1.05
  & 87.74 $\pm$ 0.91
  & $-11.71\%$
  & 0.5248 $\pm$ 0.0182
  & 94.6\% \\

  Classical Servo 3D
  & 102.12 $\pm$ 1.16
  & 99.77 $\pm$ 1.13
  & 85.97 $\pm$ 3.35
  & $-15.82\%$
  & \textbf{0.6818 $\pm$ 0.0195}
  & 84.9\% \\
  \bottomrule
  \end{tabular}
\end{table*}

Table~\ref{tab:moving_target} reveals a complementary trade-off. Classical control provides stronger nominal tracking, and the Classical Servo 3D retains the highest successful-step rate under the hard helix. However, the learned policies degrade substantially less as target dynamics depart from the training distribution. Visual Curriculum PPO shows only a $3.94\%$ reward reduction from static to helix and retains $97.7\%$ survival, compared with $15.82\%$ degradation and $84.9\%$ survival for Classical Servo 3D. The curriculum therefore provides its clearest advantage under distribution shift rather than nominal static tracking.

\subsection{Robustness to Sensing and Control Perturbations}

We additionally evaluate the action noise, visual noise, and intermittent visual dropout defined in the robustness protocol, each at perturbation level $0.2$. Rather than reporting another large table of absolute metrics, Table~\ref{tab:robustness_compact} reports reward change relative to each method's clean evaluation and survival under the most challenging visual-noise condition.

\begin{table}[t]
  \caption{Evaluation-time robustness. Action, Visual, and Dropout denote percentage reward change relative to clean evaluation; Vis. Surv. denotes survival rate under visual noise.}
  \label{tab:robustness_compact}
  \centering
  \footnotesize
  \setlength{\tabcolsep}{2.5pt}
  \begin{tabular}{lrrrr}
  \toprule
  \textbf{Method} &
  \textbf{Action} &
  \textbf{Visual} &
  \textbf{Dropout} &
  \textbf{Vis. Surv.} \\
  \midrule
  Direct PPO         & $+2.08\%$ & $-27.09\%$ & $+1.75\%$ & $97.3\%$ \\
  Visual Curr. PPO   & $+0.01\%$ & $-25.43\%$ & $+0.40\%$ & $97.2\%$ \\
  Classical Servo 2D & $-9.08\%$ & $-36.62\%$ & $+0.46\%$ & $86.7\%$ \\
  Classical Servo 3D & $-6.16\%$ & $-39.05\%$ & $-0.16\%$ & $85.1\%$ \\
  \bottomrule
  \end{tabular}
\end{table}

The PPO policies are considerably less sensitive to action noise than the classical controllers. Visual noise is the strongest perturbation for all methods, but reward degradation remains smaller for Direct PPO ($27.09\%$) and Visual Curriculum PPO ($25.43\%$) than for Classical Servo 2D ($36.62\%$) and Classical Servo 3D ($39.05\%$). The PPO policies also retain survival above $97\%$, compared with approximately $85$--$87\%$ for classical control. Dropout has little effect across methods. These results indicate that the learned policies sacrifice some nominal tracking performance but exhibit greater relative robustness to evaluation-time sensing and actuation perturbations.

\subsection{Camera-Intrinsics Generalisation}

Finally, we evaluate sensitivity to camera calibration by varying the focal length by $\pm20\%$ at evaluation time while keeping all other parameters fixed. Since the visual interface uses normalised image-space offsets rather than raw pixel coordinates or learned image features, this experiment tests whether the controllers remain functional under moderate changes in sensing geometry.

\begin{table}[t]
  \caption{Performance change under $\pm20\%$ focal-length variation relative to nominal camera intrinsics. $\Delta R$ and $\Delta S$ denote changes in reward and successful-step rate, respectively.}
  \label{tab:intrinsics}
  \centering
  \footnotesize
  \setlength{\tabcolsep}{2pt}
  \begin{tabular}{lrrrr}
  \toprule
  \textbf{Method} &
  \textbf{$\Delta R_{0.8}$} &
  \textbf{$\Delta R_{1.2}$} &
  \textbf{$\Delta S_{0.8}$} &
  \textbf{$\Delta S_{1.2}$} \\
  \midrule
  Direct PPO (Full)     & $+0.4\%$ & $-0.7\%$ & $+0.018$ & $-0.002$ \\
  Direct PPO (No Depth) & $-1.2\%$ & $+0.3\%$ & $+0.009$ & $-0.005$ \\
  Classical Servo 2D    & $-0.4\%$ & $+0.6\%$ & $+0.000$ & $+0.001$ \\
  Classical Servo 3D    & $-0.8\%$ & $+0.5\%$ & $-0.006$ & $-0.001$ \\
  \bottomrule
  \end{tabular}
\end{table}

Table~\ref{tab:intrinsics} shows limited sensitivity to focal-length changes. Across all methods and both perturbation directions, the reward change remains at or below $1.2\%$, while changes in successful-step rate are small. In particular, both PPO variants retain performance despite being evaluated with camera intrinsics not encountered during training. The classical controllers show similarly small changes, suggesting that this robustness is largely attributable to the normalised target-centric visual representation rather than to the learned policy alone. These results indicate that the compact interface provides tolerance to moderate focal-length variation, although broader changes in camera geometry remain to be investigated.

\section{Discussion}

The experiments reveal three broader observations about compact vision-guided UAV control. First, matched-budget training shows that curriculum learning is not required for PPO to solve the nominal static-target task, with Direct PPO reaching performance comparable to the curriculum variants. However, under unseen target motion, Visual Curriculum PPO exhibits the smallest relative reward degradation while retaining high survival. This suggests that visual curriculum progression can influence learned control behavior primarily under distribution shift rather than nominal evaluation.

Second, comparison with classical visual servoing reveals a trade-off between nominal accuracy and robustness. The classical controllers achieve stronger nominal tracking, particularly with explicit depth regulation, whereas the learned policies exhibit smaller degradation under action and visual perturbations and maintain higher survival under challenging shifts. Thus, their advantage is tolerance to deviations from training or tuning conditions rather than superior nominal precision.

Finally, observation ablations show that proprioception is important for stable flight and image-space cues for visual alignment, while explicit depth is less critical in the evaluated range. Together, these results indicate that observation design and training progression should be assessed not only through nominal performance, but also under sensing, dynamics, and target-motion shifts.

\section{Limitations and Future Work}

Although the framework demonstrates stable long-horizon aerial visual servoing using compact target-centric cues and low-dimensional sensor measurements, several limitations remain. The current perception pipeline uses simplified target segmentation and simulator-provided depth cues, and therefore does not address cluttered scenes, severe occlusions, multiple targets, semantic scene understanding, or real-world perception failures. However, the modular perception block can be replaced by stronger detectors, trackers, segmentation models, or depth estimators without changing the policy interface.

The study remains simulation-based and focuses on controlled analysis of the compact representation and curriculum strategies rather than exhaustive deployment conditions. While the learned policies generalise zero-shot to unseen target motion and exhibit greater relative robustness than classical controllers under sensing and control perturbations, these experiments do not capture the full perception and dynamics mismatch of physical flight. Future work will investigate realistic learned perception, prolonged target loss and reacquisition, delayed observations, broader camera and dynamics variation, and sim-to-real deployment.

\section{Conclusion}

We investigated long-horizon aerial visual servo control using compact target-centric cues and low-dimensional sensor measurements. Rather than relying on high-dimensional image encoders, the framework combines mask-derived image offsets and relative depth with quadrotor state measurements into a compact control-oriented interface. Under matched optimisation budgets, Direct PPO and the three curriculum-trained variants achieve comparable nominal performance with different tracking trade-offs. Observation ablations establish complementary roles for proprioceptive and image-space information, while explicit depth is not essential in the evaluated setting. Compared with tuned classical visual-servo controllers, learned policies sacrifice some nominal performance but exhibit greater relative robustness to sensing and control perturbations.

Policies trained only with static targets also generalise to unseen target motion without retraining, with Visual Curriculum PPO exhibiting the smallest reward degradation under the hard helical trajectory. Overall, compact target-centric representations provide an efficient interface for aerial visual servoing, while curriculum learning can shape tracking trade-offs and robustness beyond the training distribution despite comparable nominal performance.

\bibliographystyle{unsrt}
\bibliography{references}

\end{document}